\pdfoutput=1

\documentclass[11pt]{article}

\usepackage{acl}

\usepackage{times}
\usepackage{latexsym}
\usepackage{todonotes}

\usepackage[T1]{fontenc}

\usepackage[utf8]{inputenc}

\usepackage{microtype}

\usepackage[mode=buildnew]{standalone}
\usepackage{algorithm}
\usepackage{algorithmic}
\usepackage{booktabs}
\usepackage{subcaption}
\usepackage{arydshln}
\usepackage{xspace}

\usepackage{graphicx}
\usepackage{listings} 
\usepackage{enumitem}

\usepackage{amsmath,amsfonts,bm}

\def\eqref#1{equation~\ref{#1}}

\def\1{\bm{1}}

\DeclareMathAlphabet{\mathsfit}{\encodingdefault}{\sfdefault}{m}{sl}
\SetMathAlphabet{\mathsfit}{bold}{\encodingdefault}{\sfdefault}{bx}{n}

\newcommand{\spider}{\textsc{Spider}\xspace}
\newcommand{\geo}{\textsc{GeoQuery}\xspace}

\newcommand{\comment}[1]{}

\def\|#1|{\mathid{#1}}
\newcommand{\mathid}[1]{\ensuremath{\mathit{#1}}}
\def\<#1>{\codeid{#1}}
\protected\def\codeid#1{\ifmmode{\mbox{\smaller\ttfamily{#1}}}\else{\ttfamily
		#1}\fi}

\title{Learning to Synthesize Data for Semantic Parsing}

\author{
    Bailin Wang$^{\dag}$\Thanks{\ Work done at Salesforce Research. Bailin was doing a research internship.} \quad 
    Wenpeng Yin$^{\ddag}$ \quad Xi Victoria Lin$^{\mathsection}$$^*$ \and Caiming Xiong$^{\ddag}$ \\
    $^{\dag}$ University of Edinburgh, $^{\ddag}$ Salesforce Research, $^{\mathsection}$ Facebook AI \\
    {bailin.wang@ed.ac.uk}, {\{wyin,cxiong\}@salesforce.com}, {victorialin@fb.com}
}

\date{}

\begin{document}

\maketitle
\begin{abstract}
Synthesizing data for semantic parsing has gained increasing attention recently.
However, most methods require handcrafted (high-precision) rules in their generative process,
hindering the exploration of diverse unseen data.
In this work, we propose a generative model which features a (non-neural) PCFG that models
the composition of programs (e.g., SQL),  and a BART-based translation model that maps a program to an utterance. 
Due to the simplicity of PCFG and pre-trained BART, our generative model can be efficiently learned from existing data 
at hand. Moreover, explicitly modeling compositions using PCFG leads to a better
exploration of unseen programs, thus generate more diverse data.
We evaluate our method in both in-domain and out-of-domain
settings of text-to-SQL parsing on the standard benchmarks of \geo and \spider, respectively.  
Our empirical results show that the synthesized data generated from our model
can substantially help a semantic parser achieve better compositional and domain generalization.
\end{abstract}

\section{Introduction}



Recently, synthesizing data for semantic parsing has gained increasing attention~\cite{yu-etal-2018-syntaxsqlnet,yu2020grappa,zhong-etal-2020-grounded}. 
However, these models require handcrafted rules (or templates) to synthesize new programs or utterance-program
pairs. This can be sub-optimal as fixed rules cannot capture the underlying distribution of programs which
usually vary across different domains~\cite{herzig-berant-2019-dont}. Meanwhile, designing such rules also requires 
human involvement with expert knowledge. To alleviate this, we propose to learn a 
generative model from the existing data at hand.
Our key observation is that programs (e.g., SQL) are formal languages that 
are intrinsically compositional. That is, the underlying grammar of programs is usually known
and can be used to  model the space of all possible programs effectively. Typically, grammars 
are used to constrain the program space during decoding of neural
parsers~\cite{yin-neubig-2018-tranx,krishnamurthy-etal-2017-neural}.
In this work, we utilize grammars to generate (unseen) programs,
which are then used to synthesize more parallel data for semantic parsing.

\begin{figure}[t]
    \centering
    \includegraphics[width=0.45\textwidth]{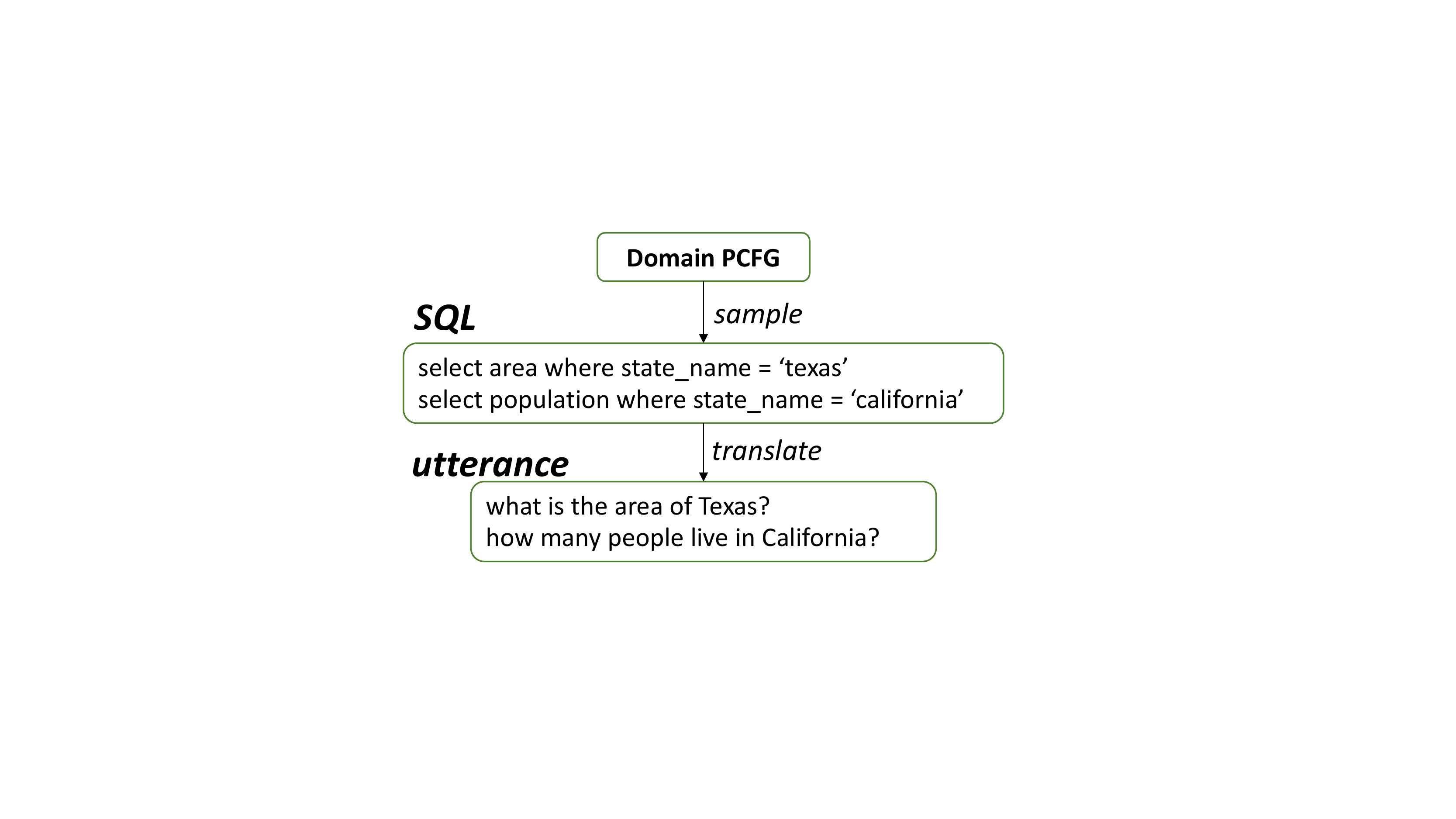} 
  \caption{A two-stage generative process for synthesizing utterance-SQL pairs.}
  \label{fig:example}
  \vspace{-5mm}
\end{figure}

Concretely, we use text-to-SQL as an example task, and propose a generative 
model to synthesize utterance-SQL pairs. As illustrated in Figure~\ref{fig:example}, 
we first employ a probabilistic context-free grammar (PCFG) to model 
the distribution of SQL queries. Then with the help of a SQL-to-text translation model, the corresponding 
utterances of SQL queries are generated subsequently. 
Our approach is in the same spirit as back-translation~\cite{sennrich-etal-2016-improving}. 
The major difference is that the `target language', in our case, is a formal language with
known underlying grammar. 
Just like the training of a semantic parser, the training of the data synthesizer 
requires a set of utterance-SQL pairs. Hence, our generative model is unlikely to be useful if it 
is as data-hungry as a semantic parser. 
Our two-stage data synthesis approach, i.e. the PCFG and the translation model, is 
designed to be more sample-efficient compared to a neural semantic parser. 
To achieve better sample efficiency,
we use the non-neural parameterization of PCFG~\cite{manning1999foundations} and estimate it via
simple counting. 
For the translation model, we use the pre-trained text generation model
BART~\cite{lewis-etal-2020-bart}.
We sample synthetic data from the generative model to pre-train 
a semantic parser. The resulting parameters can presumably
provide a strong compositional inductive bias in the form of initializations.

We conduct experiments on two text-to-SQL parsing datasets, namely \geo~\cite{zelle1996learning}
and \spider~\cite{yu-etal-2018-spider}. In the query split of \geo, where training and 
test sets do not share SQL patterns, synthesized data helps boost the performance of 
a base parser by a large margin of 12.6\%, leading to better \textit{compositional generalization} of a parser.
In the cross-domain~\footnote{We use the terms domain and database interchangeably.}
setting of \spider, synthesized data also boosts the performance
by 3.1\% in terms of execution accuracy, resulting in better 
\textit{domain generalization} of a parser.
Our work can be summarized as follows:
%
\begin{itemize}[label={$\bullet$}, topsep=1pt, itemsep=1pt]
  \item We propose to efficiently learn a generative model that can synthesize 
  parallel data for semantic parsing. 
  \item We empirically show that the synthesized data can help a neural parser 
  achieve better compositional and domain generalization.
  Our code and data are available at \url{https://github.com/berlino/tensor2struct-public}.
\end{itemize}
\section{Related Work}

\paragraph{Data Augmentation}
Data augmentation for semantic parsing has gained increasing attention
in recent years. \citet{dong-etal-2017-learning} use back-translation~\cite{sennrich-etal-2016-improving}
to obtain paraphrase of questions. \citet{jia-liang-2016-data} induce a high-precision SCFG 
from training data to generate more new ``recombinant'' examples. \citet{yu-etal-2018-syntaxsqlnet,yu2020grappa} 
follow the same spirit and use a handcrafted SCFG rule to generate new parallel data.
However, the production rules of these approaches usually have low coverage of 
meaning representations. In this work, instead of using SCFG that accounts for rigid alignments 
between utterance and programs, we use a two-stage approach that implicitly models the 
alignments by taking advantage of powerful conditional text generators such as BART. 
In this way, our approach 
can generate more diverse data.
The most related work to ours is GAZP~\cite{zhong-etal-2020-grounded} which synthesizes
parallel data directly on test databases in the context of cross-database semantic parsing.
Our work complements GAZP and shows that synthesizing data indirectly in training databases 
can also be beneficial for cross-database semantic parsing. Crucially, we learn the distribution
of SQL programs instead of relying on handcrafted templates as in GAZP. The induced distribution
helps a model explore unseen programs, leading to better compositional generalization of a parser. 

\paragraph{Generative Models}
In the history of semantic parsing, grammar-based generative models~\cite{wong-mooney-2006-learning,wong-mooney-2007-learning,zettlemoyer2012learning,lu-etal-2008-generative}
have played an important role. However, learning and inference of such models are usually expensive as they typically require grammar induction (from text to logical forms).
Moreover, their grammars are designed specifically for linguistically faithful languages, e.g., logical forms, thus not 
suitable for 
programming languages such as SQL. In contrast, our generative model is more flexible 
and efficient to train due to the two-stage decomposition. 

\section{Method}

In this section, we explain how our method can be 
applied to text-to-SQL parsing. 


\subsection{Problem Definition}

Formally, the labeled data for text-to-SQL parsing is given as 
a set of triples $(x, d, y)$, and each triple represents an
utterance $x$, the corresponding SQL query $y$ and relational database $d$.
A probabilistic semantic parser is trained to maximize $p(y|x, d)$.
The goal of this work is to learn a generative model of $q(x,y |d)$ given databases
such that it can synthesize more data (i.e., triplets) for training a semantic parser $p(y|x, d)$.
Note that we use different notations $q$ and $p$ to represent the \textit{generative model} and 
the \textit{discriminative parser}, respectively,
where $p(y|x, d)$ is not a posterior distribution of $q$. Instead, $p$ is 
a separate model trained with different parameterization with $q$. This is primarily due 
to the intractability of posterior inference of $q(y|x, d)$.
Specifically, we use a two-stage process to model the generation of utterance-SQL pairs
as follows:
%
\begin{align}
    q(x,y|d) = q(y|d) q(x|y,d)
\end{align}
where $q(y|d)$ models the distribution of SQLs given a database, and $q(x|y,d)$  
models the translation process from SQL to utterances.

\subsection{Database-Specific PCFG: $q(y|d)$}

\begin{figure}[t]
    \centering
  \begin{lstlisting}[basicstyle=\fontfamily{cmtt}\small,columns=fullflexible,frame=bt]
  sql = (select select, cond? where)
  select = (agg* aggs)
  agg = (agg_type agg_id, column col_id)
  agg_type = NoneAggOp | Max | Min 
  cond = And(cond left, cond right) 
        | Or(cond left, cond right) 
        | Not(cond c)
  \end{lstlisting}
  \caption{A simplified ASDL grammar for SQL, where ``sql, select, cond, agg" stands for variable types,
  ``where, agg\_id" for variable names, and ``And, Or, Not" for constructor names.}
  \label{fig:asdl}
  \vspace{-5mm}
\end{figure}

We use abstract syntax trees (ASTs) to model the underlying grammar of SQL,
following~\citet{yin-neubig-2018-tranx} and \citet{wang-etal-2020-rat}.
Specifically, we use ASDL~\cite{wang1997zephyr} formalism to define ASTs.
To illustrate, Figure~\ref{fig:asdl} shows a simplified ASDL grammar for SQL.
The ASDL grammar of SQL can be represented by a set of context-free grammar (CFG) rules,
as elaborated in the Appendix. 
By assuming the strong independence of each production rule, we model the 
probability of generating a SQL as the product of the probability of 
each production rule $q(y) = \prod_{i=}^N q(T_i)$.
It is well known that estimating the probability of 
a production rule via maximum-likelihood training is equivalent
to simple counting, which is defined as follows:
\vspace{-2mm}
\begin{equation}
  q(N \rightarrow \zeta) = \frac{C(N \rightarrow \zeta)}{\sum_\gamma C(N \rightarrow \gamma)}
  \label{eq:pcfg}
\end{equation}
where $C$ is the function that counts the number of occurrences of a production rule.

\subsection{SQL-to-utterance Translation: $q(x|y, d)$}

With generated SQL queries at hand, we then show how we
map SQLs to utterances to obtain more paired data.
We notice that SQL-to-utterance translation, which belongs to the general task of 
conditional text generation, shares the same output space with summarization and machine translation.
Fortunately, pre-trained models~\cite{devlin2018bert,radford2019language} using self-supervised methods have shown great
success for conditional text generation tasks. Hence, we take advantage of a contemporary pre-trained 
model, namely BART~\cite{lewis-etal-2020-bart},
which is an encoder-decoder model that uses the Transformer architecture\cite{vaswani2017attention}. 

To obtain a SQL-to-utterance translation model, we fine-tune the pre-trained BART model with our parallel data,
with SQL being the input sequence and utterance being the output sequence. 
Empirically, we found that the desired translation model can be effectively obtained using the 
SQL-utterance pairs at hand, although the original BART model is designed for text-to-text translation only.

\subsection{Semantic Parser: $p(y|x, d)$}
\label{subsec:parser}

After obtaining a trained generative model $q(x,y|d)$, we can sample synthetic pairs of $(x,y)$
for each database $d$. The synthesized data will then be used as a complement to
the original training data for a semantic parser. 
Following \citet{yu2020grappa}, we use the strategy of first pre-training a parser with the synthesized data, 
and then fine-tuning it with the original training data. In this manner, the resulting parameters
encode the compositional inductive bias introduced by our generative model.
Another way to view pre-training is that a parser $p(y|x,d)$ is essentially trained to approximate the 
posterior distribution of $q(y|x, d)$ via massive samples from $q(x,y|d)$.
\section{Experiments}

We show that our generative model can be used to
synthesize data in two settings of semantic parsing.
We also present an ablation study for our approach.

\paragraph{In-Domain Setting}

We first evaluate our method in the conventional in-domain setting 
where training and test data are from the same database.
Specifically, we synthesize new data for the \geo dataset~\cite{zelle1996learning}
which contains 880 utterance-SQL pairs on the database of U.S. geography.
We evaluate in both question and query split, following~\citet{finegan-dollak-etal-2018-improving}.
The traditional question split ensures that no utterance is repeated between the train and 
test sets. This only tests limited generalization as many utterances correspond to the same 
SQL query; query split is introduced to ensure that neither utterances nor SQL queries repeat.
The query split tests compositional generalization of a semantic parser as only fragments of test SQL
queries occur in the training set. 

\paragraph{Out-of-Domain Setting}

Then we evaluate our method in a challenging out-of-domain setting where 
the training and test databases do not overlap. That is,
a parser is trained on some \textit{source databases} but evaluated in unseen \textit{target databases}.
Concretely, we apply our method to the \spider~\cite{yu-etal-2018-spider} dataset 
where the training contains utterance-SQL pairs from 146 source databases and the test set contains
data from a disjoint set of target databases.
In this out-of-domain setting, we synthesize data in the source databases in the hope 
that it can promote its domain generalization to unseen target databases.

\paragraph{Training}

\begin{table}[t!]
    \scalebox{0.72}{
        \begin{tabular}{l|cc}
            \toprule
            Model &  Question Split  & Query Split \\
            \midrule
            seq2tree \citep{dong-lapata-2016-language} & 62 \rlap{$^\dagger$} & 31 \rlap{$^\dagger$} \\
            GECA \citep{andreas-2020-good} & 68 \rlap{$^\dagger$} & 49 \rlap{$^\dagger$} \\
            template-based \shortcite{finegan-dollak-etal-2018-improving}  & 55.2  & - \\
            seq2seq \citep{iyer-etal-2017-learning}   &  72.5 & - \\
            \hdashline
            Base Parser$^\clubsuit$&    70.9  & 49.5  \\
            Base Parser$^\clubsuit$ + Syn Pre-Train & \bf 74.6 & \bf 62.1 \\
            \quad  $w.o.$ trained PCFG &   72.4 & 54.8 \\ 
            \quad  $w.o.$ pre-trained BART &  71.5 & 53.9 \\ 
            \bottomrule
        \end{tabular}
    }
    \caption{Execution accuracies on \geo. Methods with $^\dagger$ measure exact match accuracy.
    $w.o.$ stands for ablating a certain component.}
    \label{tab:main_results_geo}
\end{table}

As mentioned in Section~\ref{subsec:parser}, we use pre-training to 
augment a semantic parser with synthesized data.
Specifically, we use the following four-step training procedure:
1) train a two-stage generative model, namely $q(x,y|d)$, 2) sample 
new data from it, 3) pre-train a semantic parser $p(y|x,d)$ 
using the synthesized data, 4) fine-tune the parser with the target training data. 
In the in-domain setting, one PCFG and translation model is trained.
In the out-of-domain setting, a separate PCFG is trained on each source database assuming that 
each database has a different distribution of SQL queries. 
In contrast, a single translation model is trained and shared across source databases. 
We use RAT-SQL~\cite{wang-etal-2020-rat} as our base parser.

The size of the synthesized data is always proportional to the size of the original data.
We tune the ratio in $\{1, 3, 6, 12\}$, and find that $3$, $6$ works best
for \geo and \spider respectively. 
We use the RAT-SQL implementation from~\citet{wang2020meta} 
which supports value prediction and evaluation by execution.
We train it with the default hyper-parameters.
For the SQL-to-utterance translation model, we reuse all the default 
hyperparameters from BART~\cite{lewis-etal-2020-bart}. 
Both models are trained using NVIDIA V100. 

\subsection{Main Results}
 
\begin{table}[t!]
    \scalebox{0.72}{
        \begin{tabular}{l|cc}
            \toprule
            Model & \textsc{Set Match} & \textsc{Execution}  \\
            \midrule
            RAT-SQL$^\spadesuit$ \citep{wang-etal-2020-rat} & 69.7 & -   \\
            RYANSQL$^\spadesuit$ \citep{choi2020ryansql} & 70.6 & -    \\
            IRNet$^\diamondsuit$ \citep{guo-etal-2019-towards} & 61.9 & -  \\
            GAZP \citep{zhong-etal-2020-grounded} & 59.1 & 59.2   \\
            BRIDGE$^\spadesuit$ \citep{DBLP:journals/corr/abs-2012-12627} & 70.0 & 68.0  \\
            \hline
            Base Parser$^\clubsuit$ &  70.4 & 69.4    \\
            Base Parser$^\clubsuit$ + Syn Pre-Train & \bf 71.8 & \bf 72.5  \\
            \quad  $\emph{w.o.}$ trained PCFG & 71.4 & 72.3 \\ 
            \quad  $\emph{w.o.}$ pre-trained BART & 70.6 & 70.8  \\ 
            \bottomrule
        \end{tabular}
    }
    \caption{Set match and execution accuracies on \spider. $^\spadesuit$ stands for models with BERT-large, $^\diamondsuit$ for BERT-base,
    $^\clubsuit$ for Electra-base.  
    }
    \label{tab:main_results_spider}
\end{table}

For \geo, we report execution accuracy on the test sets of the question and query split; 
for \spider, we report exact set match~\cite{yu-etal-2018-spider} along with
execution accuracy on the dev set.  The main results are shown in Table~\ref{tab:main_results_geo} and
\ref{tab:main_results_spider}.
First, we can see that compared with previous work, our base parser achieves the best performance, 
confirming that we are using a strong base parser to test our synthesized data.

With the pre-training using synthesized data, the performance of the base parsers is boosted in
both \geo and \spider. In \geo, the pre-training results in the margin of 12.6\% in the query split.
This is somewhat expected as our generative model, especially $q(y|d)$ directly models the composition 
underlying SQL queries, which helps a parser generalize better to unseen queries. 
Moreover, our sampled SQL queries cover around 15\% test SQL queries of the query split, partially explaining 
why it is so beneficial for the query split. 
In \spider, the pre-training boosts the performance by 3.1\% in terms of execution accuracy.
Although our model does not synthesize data directly for target databases (which are unseen), it still helps a parser achieve 
better domain generalization. This contradicts the observation by~\citet{zhong-etal-2020-grounded}
that synthesizing data in source databases is useless, even harmful without careful consistency calibration. We attribute this to the pre-training strategy
we use, as in our preliminary experiments we found that directly mixing the synthesized data with the original training data is indeed harmful.

\subsection{Ablation Study}

\begin{table*}[t]
    \centering
    \scalebox{0.82}{
    \begin{tabular}{@{}ll@{}}
      \toprule
       \textbf{Sampled SQLs ($y$)}
       & \textbf{Generated Utterances ($x$)} \\
      \midrule
       {\small SELECT length FROM river WHERE traverse = "new york"}
       & {\small What is the length of the river whose traverse is in New York city?} \\
       {\small SELECT Sum(length) FROM river WHERE traverse = "colorado"}
       & {\small What is the total length of the rivers that traverse the state of Colorado?} \\
       {\small SELECT state\_name FROM border\_info WHERE border = "wyoming"}
       & {\small What are the names of the states that have a border with Wyoming?} \\
      \midrule
       {\small SELECT state\_name FROM city WHERE population = "mississippi"}
       & {\small What are the names of all cities in the state of Mississippi?} \\
       {\small SELECT Min(state\_name) FROM state WHERE state\_name = "mississippi" }
       & {\small What is the minimum state name of the state with the name Mississippi?} \\
       {\small SELECT capital FROM state WHERE population = 15000}
       & {\small What are the capitals of states with population of 150000 or more?} \\
      \bottomrule
    \end{tabular}}
    \caption{Positive (top) and negative (bottom) examples of synthesized paired data for \geo.}
  \label{tab:qualitative}
  \end{table*}

We try to answer two questions: a) \textit{whether it is necessary to 
learn a PCFG}; b) \textit{whether pre-trained translation model, namely BART, 
is required for success}. To answer the first question,
we use a randomized version of $q(y|d)$ where the probability of
production rules are uniformly distributed, instead of being estimated
from data in Equation~(\ref{eq:pcfg}). As shown in Table~\ref{tab:main_results_geo} and \ref{tab:main_results_spider},
this variant ($\emph{w.o.}$ trained PCFG) still improves the base parsers, but with a smaller margin. 
This shows that a trained PCFG model is better at synthesizing useful SQL queries.
To answer the second question, we use a randomly initialized SQL-to-utterance translation model 
instead of BART. As shown in Table~\ref{tab:main_results_geo} and \ref{tab:main_results_spider},
this variant ($\emph{w.o.}$  pre-trained BART) results in a drop in performance as well,
indicating that pre-trained BART is crucial for synthesizing useful utterances.

\subsection{Qualitative Analysis}

Table~\ref{tab:qualitative} shows examples of synthesized paired data 
for \geo. In the positive examples, the sampled SQLs can be viewed as recombinations
of SQLs fragments observed in the training data. For example, \<SELECT Sum(length)> 
and \<traverse = ``colorado''> are SQL fragments from separate training examples. Our PCFG 
combines them together to form a new SQL, and the SQL-to-utterance model successfully 
maps it to a reasonable translation. The negative examples consist of two kinds of 
errors. First, the PCFG generated semantically invalid SQLs which cannot be mapped 
to reasonable utterances. This error is due to the independence assumption made by the PCFG.
For instance, when a column and its corresponding entity is separately sampled, there is no
guarantee that they form a meaningful clause, as shown in \<population = ``mississippi''>. 
To address this, future work might consider more powerful 
generative models to model the dependencies within and across clauses in a SQL.
Second, the SQL-to-utterances model failed to translate the sampled SQLs, as shown 
in the last example. 

\section{Conclusion}

In this work, we propose to efficiently learn a generative model 
that can synthesize parallel data for semantic parsing.
The synthesized data is used to pre-train a semantic parser 
and provide a strong inductive bias of compositionality. Empirical 
results on \geo and \spider show that the pre-training can help 
a parser achieve better compositional and domain generalization.

\section*{Acknowledgements}
We would like to thank the anonymous reviewers for their valuable comments.
We thank Naihao Deng for providing the preprocessed database for \geo.

\bibliography{bib/anthology,bib/rebibed_main,bib/text2sql-datasets}

\begin{thebibliography}{30}
\expandafter\ifx\csname natexlab\endcsname\relax\def\natexlab#1{#1}\fi

\bibitem[{Andreas(2020)}]{andreas-2020-good}
Jacob Andreas. 2020.
\newblock \href {https://doi.org/10.18653/v1/2020.acl-main.676} {Good-enough
  compositional data augmentation}.
\newblock In \emph{Proceedings of the 58th Annual Meeting of the Association
  for Computational Linguistics}, pages 7556--7566, Online. Association for
  Computational Linguistics.

\bibitem[{Choi et~al.(2020)Choi, Shin, Kim, and Shin}]{choi2020ryansql}
DongHyun Choi, Myeong~Cheol Shin, EungGyun Kim, and Dong~Ryeol Shin. 2020.
\newblock Ryansql: Recursively applying sketch-based slot fillings for complex
  text-to-sql in cross-domain databases.
\newblock \emph{arXiv preprint arXiv:2004.03125}.

\bibitem[{Devlin et~al.(2019)Devlin, Chang, Lee, and
  Toutanova}]{devlin2018bert}
Jacob Devlin, Ming-Wei Chang, Kenton Lee, and Kristina Toutanova. 2019.
\newblock \href {https://doi.org/10.18653/v1/N19-1423} {{BERT}: Pre-training of
  deep bidirectional transformers for language understanding}.
\newblock In \emph{Proceedings of the 2019 Conference of the North {A}merican
  Chapter of the Association for Computational Linguistics: Human Language
  Technologies, Volume 1 (Long and Short Papers)}, pages 4171--4186,
  Minneapolis, Minnesota. Association for Computational Linguistics.

\bibitem[{Dong and Lapata(2016)}]{dong-lapata-2016-language}
Li~Dong and Mirella Lapata. 2016.
\newblock \href {https://doi.org/10.18653/v1/P16-1004} {Language to logical
  form with neural attention}.
\newblock In \emph{Proceedings of the 54th Annual Meeting of the Association
  for Computational Linguistics (Volume 1: Long Papers)}, pages 33--43, Berlin,
  Germany. Association for Computational Linguistics.

\bibitem[{Dong et~al.(2017)Dong, Mallinson, Reddy, and
  Lapata}]{dong-etal-2017-learning}
Li~Dong, Jonathan Mallinson, Siva Reddy, and Mirella Lapata. 2017.
\newblock \href {https://doi.org/10.18653/v1/D17-1091} {Learning to paraphrase
  for question answering}.
\newblock In \emph{Proceedings of the 2017 Conference on Empirical Methods in
  Natural Language Processing}, pages 875--886, Copenhagen, Denmark.
  Association for Computational Linguistics.

\bibitem[{Finegan-Dollak et~al.(2018)Finegan-Dollak, Kummerfeld, Zhang,
  Ramanathan, Sadasivam, Zhang, and Radev}]{finegan-dollak-etal-2018-improving}
Catherine Finegan-Dollak, Jonathan~K. Kummerfeld, Li~Zhang, Karthik Ramanathan,
  Sesh Sadasivam, Rui Zhang, and Dragomir Radev. 2018.
\newblock \href {https://doi.org/10.18653/v1/P18-1033} {Improving text-to-{SQL}
  evaluation methodology}.
\newblock In \emph{Proceedings of the 56th Annual Meeting of the Association
  for Computational Linguistics (Volume 1: Long Papers)}, pages 351--360,
  Melbourne, Australia. Association for Computational Linguistics.

\bibitem[{Guo et~al.(2019)Guo, Zhan, Gao, Xiao, Lou, Liu, and
  Zhang}]{guo-etal-2019-towards}
Jiaqi Guo, Zecheng Zhan, Yan Gao, Yan Xiao, Jian-Guang Lou, Ting Liu, and
  Dongmei Zhang. 2019.
\newblock \href {https://doi.org/10.18653/v1/P19-1444} {Towards complex
  text-to-{SQL} in cross-domain database with intermediate representation}.
\newblock In \emph{Proceedings of the 57th Annual Meeting of the Association
  for Computational Linguistics}, pages 4524--4535, Florence, Italy.
  Association for Computational Linguistics.

\bibitem[{Herzig and Berant(2019)}]{herzig-berant-2019-dont}
Jonathan Herzig and Jonathan Berant. 2019.
\newblock \href {https://doi.org/10.18653/v1/D19-1394} {Don{'}t paraphrase,
  detect! rapid and effective data collection for semantic parsing}.
\newblock In \emph{Proceedings of the 2019 Conference on Empirical Methods in
  Natural Language Processing and the 9th International Joint Conference on
  Natural Language Processing (EMNLP-IJCNLP)}, pages 3810--3820, Hong Kong,
  China. Association for Computational Linguistics.

\bibitem[{Iyer et~al.(2017)Iyer, Konstas, Cheung, Krishnamurthy, and
  Zettlemoyer}]{iyer-etal-2017-learning}
Srinivasan Iyer, Ioannis Konstas, Alvin Cheung, Jayant Krishnamurthy, and Luke
  Zettlemoyer. 2017.
\newblock \href {https://doi.org/10.18653/v1/P17-1089} {Learning a neural
  semantic parser from user feedback}.
\newblock In \emph{Proceedings of the 55th Annual Meeting of the Association
  for Computational Linguistics (Volume 1: Long Papers)}, pages 963--973,
  Vancouver, Canada. Association for Computational Linguistics.

\bibitem[{Jia and Liang(2016)}]{jia-liang-2016-data}
Robin Jia and Percy Liang. 2016.
\newblock \href {https://doi.org/10.18653/v1/P16-1002} {Data recombination for
  neural semantic parsing}.
\newblock In \emph{Proceedings of the 54th Annual Meeting of the Association
  for Computational Linguistics (Volume 1: Long Papers)}, pages 12--22, Berlin,
  Germany. Association for Computational Linguistics.

\bibitem[{Krishnamurthy et~al.(2017)Krishnamurthy, Dasigi, and
  Gardner}]{krishnamurthy-etal-2017-neural}
Jayant Krishnamurthy, Pradeep Dasigi, and Matt Gardner. 2017.
\newblock \href {https://doi.org/10.18653/v1/D17-1160} {Neural semantic parsing
  with type constraints for semi-structured tables}.
\newblock In \emph{Proceedings of the 2017 Conference on Empirical Methods in
  Natural Language Processing}, pages 1516--1526, Copenhagen, Denmark.
  Association for Computational Linguistics.

\bibitem[{Lewis et~al.(2020)Lewis, Liu, Goyal, Ghazvininejad, Mohamed, Levy,
  Stoyanov, and Zettlemoyer}]{lewis-etal-2020-bart}
Mike Lewis, Yinhan Liu, Naman Goyal, Marjan Ghazvininejad, Abdelrahman Mohamed,
  Omer Levy, Veselin Stoyanov, and Luke Zettlemoyer. 2020.
\newblock \href {https://doi.org/10.18653/v1/2020.acl-main.703} {{BART}:
  Denoising sequence-to-sequence pre-training for natural language generation,
  translation, and comprehension}.
\newblock In \emph{Proceedings of the 58th Annual Meeting of the Association
  for Computational Linguistics}, pages 7871--7880, Online. Association for
  Computational Linguistics.

\bibitem[{Lin et~al.(2020)Lin, Socher, and
  Xiong}]{DBLP:journals/corr/abs-2012-12627}
Xi~Victoria Lin, Richard Socher, and Caiming Xiong. 2020.
\newblock \href {http://arxiv.org/abs/2012.12627} {Bridging textual and tabular
  data for cross-domain text-to-sql semantic parsing}.
\newblock \emph{CoRR}, abs/2012.12627.

\bibitem[{Lu et~al.(2008)Lu, Ng, Lee, and
  Zettlemoyer}]{lu-etal-2008-generative}
Wei Lu, Hwee~Tou Ng, Wee~Sun Lee, and Luke~S. Zettlemoyer. 2008.
\newblock \href {https://www.aclweb.org/anthology/D08-1082} {A generative model
  for parsing natural language to meaning representations}.
\newblock In \emph{Proceedings of the 2008 Conference on Empirical Methods in
  Natural Language Processing}, pages 783--792, Honolulu, Hawaii. Association
  for Computational Linguistics.

\bibitem[{Manning and Sch\"{u}tze(1999)}]{manning1999foundations}
Christopher Manning and Hinrich Sch\"{u}tze. 1999.
\newblock \emph{Foundations of statistical natural language processing}.
\newblock MIT press.

\bibitem[{Radford et~al.(2019)Radford, Wu, Child, Luan, Amodei, and
  Sutskever}]{radford2019language}
Alec Radford, Jeffrey Wu, Rewon Child, David Luan, Dario Amodei, and Ilya
  Sutskever. 2019.
\newblock Language models are unsupervised multitask learners.

\bibitem[{Sennrich et~al.(2016)Sennrich, Haddow, and
  Birch}]{sennrich-etal-2016-improving}
Rico Sennrich, Barry Haddow, and Alexandra Birch. 2016.
\newblock \href {https://doi.org/10.18653/v1/P16-1009} {Improving neural
  machine translation models with monolingual data}.
\newblock In \emph{Proceedings of the 54th Annual Meeting of the Association
  for Computational Linguistics (Volume 1: Long Papers)}, pages 86--96, Berlin,
  Germany. Association for Computational Linguistics.

\bibitem[{Vaswani et~al.(2017)Vaswani, Shazeer, Parmar, Uszkoreit, Jones,
  Gomez, Kaiser, and Polosukhin}]{vaswani2017attention}
Ashish Vaswani, Noam Shazeer, Niki Parmar, Jakob Uszkoreit, Llion Jones,
  Aidan~N. Gomez, Lukasz Kaiser, and Illia Polosukhin. 2017.
\newblock \href
  {https://proceedings.neurips.cc/paper/2017/hash/3f5ee243547dee91fbd053c1c4a845aa-Abstract.html}
  {Attention is all you need}.
\newblock In \emph{Advances in Neural Information Processing Systems 30: Annual
  Conference on Neural Information Processing Systems 2017, December 4-9, 2017,
  Long Beach, CA, {USA}}, pages 5998--6008.

\bibitem[{Wang et~al.(2020{\natexlab{a}})Wang, Lapata, and
  Titov}]{wang2020meta}
Bailin Wang, Mirella Lapata, and Ivan Titov. 2020{\natexlab{a}}.
\newblock Meta-learning for domain generalization in semantic parsing.
\newblock \emph{arXiv preprint arXiv:2010.11988}.

\bibitem[{Wang et~al.(2020{\natexlab{b}})Wang, Shin, Liu, Polozov, and
  Richardson}]{wang-etal-2020-rat}
Bailin Wang, Richard Shin, Xiaodong Liu, Oleksandr Polozov, and Matthew
  Richardson. 2020{\natexlab{b}}.
\newblock \href {https://doi.org/10.18653/v1/2020.acl-main.677} {{RAT-SQL}:
  Relation-aware schema encoding and linking for text-to-{SQL} parsers}.
\newblock In \emph{Proceedings of the 58th Annual Meeting of the Association
  for Computational Linguistics}, pages 7567--7578, Online. Association for
  Computational Linguistics.

\bibitem[{Wang et~al.(1997)Wang, Appel, Korn, and Serra}]{wang1997zephyr}
Daniel~C Wang, Andrew~W Appel, Jeffrey~L Korn, and Christopher~S Serra. 1997.
\newblock The zephyr abstract syntax description language.

\bibitem[{Wong and Mooney(2006)}]{wong-mooney-2006-learning}
Yuk~Wah Wong and Raymond Mooney. 2006.
\newblock \href {https://www.aclweb.org/anthology/N06-1056} {Learning for
  semantic parsing with statistical machine translation}.
\newblock In \emph{Proceedings of the Human Language Technology Conference of
  the {NAACL}, Main Conference}, pages 439--446, New York City, USA.
  Association for Computational Linguistics.

\bibitem[{Wong and Mooney(2007)}]{wong-mooney-2007-learning}
Yuk~Wah Wong and Raymond Mooney. 2007.
\newblock \href {https://www.aclweb.org/anthology/P07-1121} {Learning
  synchronous grammars for semantic parsing with lambda calculus}.
\newblock In \emph{Proceedings of the 45th Annual Meeting of the Association of
  Computational Linguistics}, pages 960--967, Prague, Czech Republic.
  Association for Computational Linguistics.

\bibitem[{Yin and Neubig(2018)}]{yin-neubig-2018-tranx}
Pengcheng Yin and Graham Neubig. 2018.
\newblock \href {https://doi.org/10.18653/v1/D18-2002} {{TRANX}: A
  transition-based neural abstract syntax parser for semantic parsing and code
  generation}.
\newblock In \emph{Proceedings of the 2018 Conference on Empirical Methods in
  Natural Language Processing: System Demonstrations}, pages 7--12, Brussels,
  Belgium. Association for Computational Linguistics.

\bibitem[{Yu et~al.(2020)Yu, Wu, Lin, Wang, Tan, Yang, Radev, Socher, and
  Xiong}]{yu2020grappa}
Tao Yu, Chien-Sheng Wu, Xi~Victoria Lin, Bailin Wang, Yi~Chern Tan, Xinyi Yang,
  Dragomir Radev, Richard Socher, and Caiming Xiong. 2020.
\newblock \href {http://arxiv.org/abs/2009.13845} {Grappa: Grammar-augmented
  pre-training for table semantic parsing}.

\bibitem[{Yu et~al.(2018{\natexlab{a}})Yu, Yasunaga, Yang, Zhang, Wang, Li, and
  Radev}]{yu-etal-2018-syntaxsqlnet}
Tao Yu, Michihiro Yasunaga, Kai Yang, Rui Zhang, Dongxu Wang, Zifan Li, and
  Dragomir Radev. 2018{\natexlab{a}}.
\newblock \href {https://doi.org/10.18653/v1/D18-1193} {{S}yntax{SQLN}et:
  Syntax tree networks for complex and cross-domain text-to-{SQL} task}.
\newblock In \emph{Proceedings of the 2018 Conference on Empirical Methods in
  Natural Language Processing}, pages 1653--1663, Brussels, Belgium.
  Association for Computational Linguistics.

\bibitem[{Yu et~al.(2018{\natexlab{b}})Yu, Zhang, Yang, Yasunaga, Wang, Li, Ma,
  Li, Yao, Roman, Zhang, and Radev}]{yu-etal-2018-spider}
Tao Yu, Rui Zhang, Kai Yang, Michihiro Yasunaga, Dongxu Wang, Zifan Li, James
  Ma, Irene Li, Qingning Yao, Shanelle Roman, Zilin Zhang, and Dragomir Radev.
  2018{\natexlab{b}}.
\newblock \href {https://doi.org/10.18653/v1/D18-1425} {{S}pider: A large-scale
  human-labeled dataset for complex and cross-domain semantic parsing and
  text-to-{SQL} task}.
\newblock In \emph{Proceedings of the 2018 Conference on Empirical Methods in
  Natural Language Processing}, pages 3911--3921, Brussels, Belgium.
  Association for Computational Linguistics.

\bibitem[{Zelle and Mooney(1996)}]{zelle1996learning}
John~M Zelle and Raymond~J Mooney. 1996.
\newblock Learning to parse database queries using inductive logic programming.
\newblock In \emph{Proceedings of the national conference on artificial
  intelligence}, pages 1050--1055.

\bibitem[{Zettlemoyer and Collins(2005)}]{zettlemoyer2012learning}
Luke~S. Zettlemoyer and Michael Collins. 2005.
\newblock Learning to map sentences to logical form: Structured classification
  with probabilistic categorial grammars.
\newblock In \emph{Proceedings of the Twenty-First Conference on Uncertainty in
  Artificial Intelligence}, UAI'05, page 658–666, Arlington, Virginia, USA.
  AUAI Press.

\bibitem[{Zhong et~al.(2020)Zhong, Lewis, Wang, and
  Zettlemoyer}]{zhong-etal-2020-grounded}
Victor Zhong, Mike Lewis, Sida~I. Wang, and Luke Zettlemoyer. 2020.
\newblock \href {https://www.aclweb.org/anthology/2020.emnlp-main.558}
  {Grounded adaptation for zero-shot executable semantic parsing}.
\newblock In \emph{Proceedings of the 2020 Conference on Empirical Methods in
  Natural Language Processing (EMNLP)}, pages 6869--6882, Online. Association
  for Computational Linguistics.

\end{thebibliography}
\bibliographystyle{acl_natbib}

\appendix
\section{CFG Rules}

Following \citet{yin-neubig-2018-tranx}, we represent ASDL grammar of SQLs 
using a set of production rules, as illustrated in Figure~\ref{fig:cfg}.
%
\lstset{
  mathescape,         
  literate={->}{$\rightarrow$ }{2}
           {;}{;   }{1} 
}
\begin{figure}[h]
    \centering
  \begin{lstlisting}[basicstyle=\fontfamily{cmtt}\small,columns=fullflexible,frame=bt]
  sql -> select;         sql -> select, cond;
  select -> agg;         select -> agg, agg;
  agg -> agg_type, column;
  agg_type -> NoneAggOp;       
  agg_type -> Min; agg_type -> Max;
  cond -> And;      cond -> Or; cond -> Not;
  \end{lstlisting}
    \caption{Context-free grammars that represent the ASDL grammar in Figure~2 of the main paper.
    Only variable types are used in the production rules.}
  \label{fig:cfg}
\end{figure}

Formally, a production rule $T$ is denoted as $N \rightarrow \zeta$, where
$N$ represents a non-terminal variable type, $\zeta$ represents a sequence 
of terminal or non-terminals. We can derive a set of production rules
from our pre-defined ASDL grammar by instantiating original ASDL statements.
For example, ``sql = (select select, cond? where)" is instantiated into two rules:
``sql $\rightarrow$ select"  and ``sql $\rightarrow$ select, cond".
With pre-defined production rules, a SQL can be transformed into 
a sequence of production rules. For example, the SQL query ``select max(age)''
can be represented by the sequence: 
\begin{enumerate}[label={(\arabic*)}]
  \setlength\itemsep{-0.5em}
  \item sql $\rightarrow$ select 
  \item select $\rightarrow$ agg
  \item agg $\rightarrow$ agg\_type, column
  \item agg\_type $\rightarrow$ Max
  \item column $\rightarrow$ age
\end{enumerate}

\end{document}